# Principal Component Analysis-Based Terahertz Self-Supervised Denoising and Deblurring Deep Neural Networks

Pengfei Zhu, *Member, IEEE*, Stefano Sfarra, Hai Zhang, *Member, IEEE*, Carlo Santulli, Elana Pivarciova, Fabrizio Sarasini, Xavier Maldague, *Life Senior Member, IEEE*

*Abstract*—Terahertz (THz) systems inherently introduce frequency-dependent degradation effects, resulting in low-frequency blurring and high-frequency noise in amplitude images. Conventional image processing techniques cannot simultaneously address both issues, and manual intervention is often required due to the unknown boundary between denoising and deblurring. To tackle this challenge, we propose a principal component analysis (PCA)-based THz self-supervised denoising and deblurring network (THz-SSDD). The network employs a Recorrupted-to-Recorrupted self-supervised learning strategy to capture the intrinsic features of noise by exploiting invariance under repeated corruption. PCA decomposition and reconstruction are then applied to restore images across both low and high frequencies. The performance of the THz-SSDD network was evaluated on four types of samples. Training requires only a small set of unlabeled noisy images, and testing across samples with different material properties and measurement modes demonstrates effective denoising and deblurring. Quantitative analysis further validates the network's feasibility, showing improvements in image quality while preserving the physical characteristics of the original signals.

## I. INTRODUCTION

WITH the non-ionizing, non-invasive, high penetration, and spectral fingerprinting features of terahertz (THz) wave, THz spectroscopy has great potential for the qualitative and quantitative identification of key substances in industrial fields [1], [2], [3]. Despite the fact that THz imaging offers a unique means to probe material's structures and properties [4], [5], it poses major challenges in terms of data processing because of the multiple sources of image degradation that make difficult the analysis of the THz signals [6].

The most signal processing research was focused on 1D THz signals because the reflected THz wave can carry the depth information of the inner boundaries. This property makes reflected THz pulses ideal for imaging multi-layer structures and bulks with inside voids and defects [7], [8]. The reflected time-domain signal comprises multiple time-shifted and amplitude-attenuated pulses, which are formed by the temporal waveform of the incident pulse convoluted with the impulse responses of the interfaces [9], [10], therefore the impulse-response function of the system can be estimated by deconvoluting the time signal by the temporal waveform of the incident pulse. However, the echoes reflected by the interfaces overlap if the layer or defect is too thin, and the echoes may submerge by noise in poor detection.

Numerical deconvolution method was first investigated to map the distribution of layers for single-layer and multi-layer paint films [11]. A method based on multiple regression analysis was later proposed to improve the resolution at the expense of signal-to-noise ratio (SNR) [12]. Researchers attempted to realize higher resolution of thickness measurement by compressing the pulse width [13]. However, the physical limit of the pump source is constrained. Subsequently, researchers focused on filtering out the noise spikes in the higher frequency region of the impulse-response function by thresholding [14]. Further study was employed to signify the impulse response by combining Wiener deconvolution and wavelet shrinkage [15]. This approach was named hybrid

This work was supported in part by the Adolf Martens Fellowship under Grant BAM-AMF-2025-1, in part by the Natural Sciences and Engineering Research Council of Canada (NSERC) through the CREATE-oN DuTy! Program under Grant 496439-2017, in part by the Canada Research Chair in Multi-polar Infrared Vision (MIVIM). (Corresponding author: Pengfei Zhu).

Pengfei Zhu is with the Division Thermographic Methods, Department of Nondestructive Testing, Bundesanstalt für Materialforschung und prüfung, 12200 Berlin, Germany (e-mail: pengfei.zhu@bam.de).

Stefano Sfarra is with the Department of Industrial and Information Engineering and Economics (DIIIE), University of L'Aquila, I-67100 L'Aquila, Italy (email: stefano.sfarra@univaq.it).

Hai Zhang is with the Department of Electrical and Computer Engineering, Computer Vision and Systems Laboratory (CVSL), Laval University, Québec G1V 0A6, Québec city, Canada, and is also with the Centre for Composite Materials and Structures (CCMS), Harbin Institute of Technology, Harbin 150001, China (email: hai.zhang.1@ulaval.ca).

Elena Pivarciova is with the Department of Manufacturing and Automation Technology, Technical University in Zvolen, 960 01, Zvolen, Slovakia (email: pivarciova@tuzvo.sk).

Carlo Santulli is with the Geology Division, School of Science and Technology (SST), Università degli Studi di Camerino, Camerino, Italy (email: carlo.santulli@unicam.it).

Fabrizio Sarasini is with the Department of Chemical Engineering Materials Environment & UDR INSTM, Sapienza University of Rome, Rome, I-00184, Italy (email: fabrizio.sarasini@uniroma1.it).

Xavier Maldague is with the Department of Electrical and Computer Engineering, Computer Vision and Systems Laboratory (CVSL), Laval University, Québec G1V 0A6, Québec city, Canada (e-mail: xavier.maldague@gel.ulaval.ca).

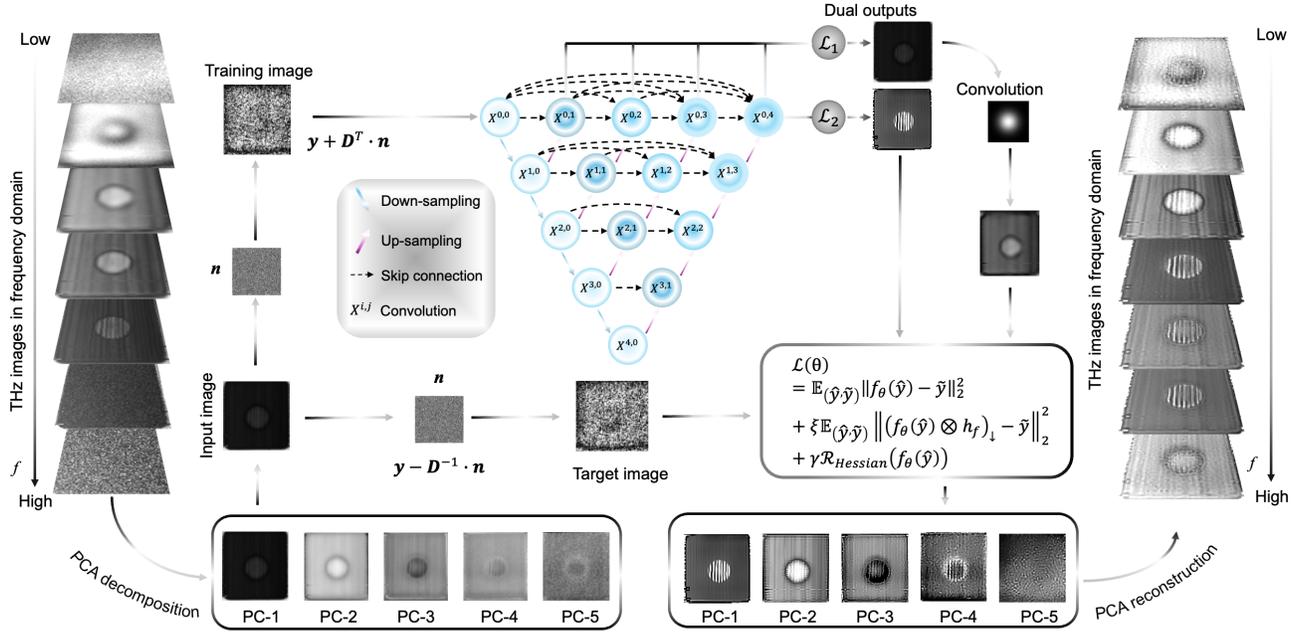

Fig. 1. **The schematic image of PCA-based THz-SSDD neural networks.** Original THz signals are transferred into frequency domain. Then, they undergo the principal component decomposition. The first five principal components are selected and further processed by the proposed THz-SSDD network. Finally, these principal components are used to reconstruct the THz amplitude spectrum.

frequency-wavelet domain deconvolution (FWDD). However, the impulse responses retrieved by FWDD are broadened by the cutoff of the low-SNR region, which limits its resolvability for overlapped echoes [16].

The design of the THz system, including the focusing optics and beam-forming process, introduces frequency-dependent degradation effects that distort the amplitude images. The primary factors contributing to the degradation and subsequent loss of spatial information in THz images include: 1) the shape of THz beam; 2) system-dependent noise; 3) losses resulting from THz wave propagation through the sample, such as reflection and/or refraction; 4) the scattering and absorption properties of the sample. While the latter two factors are inherently sample-dependent and difficult to model accurately, the effects of the THz beam shape and system noise can be estimated using simple approximations.

There are few literatures regarding 2D THz image processing. Zhu et al. [17] proposed a normalized time-domain integration method to extract important spectral information. Xu et al. [18] applied projection onto a convex sets approach, iterative backprojection approach, Lucy-Richardson iteration, and 2D wavelet decomposition reconstruction to THz imaging for obtaining a high-resolution THz image with decreased noise. Ljubenović et al. [19] proposed a joint deblurring and denoising approach for restoring high-resolution hyper-spectral images in THz-TDS.

Deep learning techniques [20], [21] have received much attention in the area of image denoising and deblurring. Jung et al. [22] proposed a self-supervised deep-learning for efficiently denoising terahertz images in terahertz time-domain spectroscopy systems. Dutta et al. [23] proposed an unsupervised deep learning network for nondestructive historical document analysis. Zhu et al. [1] proposed the THz super-resolution generative adversarial networks, which extremely increases the image quality by comparing with Richardson-Lucy algorithm. As can be seen, denoising and deblurring problem were separated. Thus, researchers always use two method / networks to mitigate these two problems.

In this study, we propose the principal component analysis (PCA)-based terahertz self-supervised denoising and deblurring deep neural network (THz-SSDD), which is a self-supervised learning framework based on the *Recorrupted-to-Recorrupted* (R2R) strategy for denoising and deconvolution. The model is trained exclusively on terahertz images from calibration plates – glass fibre reinforced polymer (GFRP) laminate with six holes. To validate the generalizability of THz-SSDD, we evaluate it on several representative samples from different industrial scenarios, including pyrolyzed woods, high-density polyethylene (HDPE) after tensile, and hybrid composite materials with impact damage. Our approach effectively addresses key limitations of existing methods, including the low robustness of conventional algorithms, the data dependency of supervised networks, and the limited generalization of current unsupervised approaches. To the best of our knowledge, this is the first attempt in terahertz time-domain spectroscopy (THz-TDS) by using only single algorithm/network to simultaneously address the noise problem at low-frequency and the blurring problem at high-frequency.

## II. METHODOLOGY

The principal component analysis (PCA)-based THz-SSDD was used to address two problems in THz-TDS, including the noise effects at high frequency and the blur effects at low frequency.

*A. Modeling of Noise*

The observed THz image $I_{obs}$ at each frequency can be modeled as the convolution of the clean signal $I$ with the frequency-dependent point spread function $h_f$, contaminated by multiple additive noise sources:

$$I_{obs} = (I \otimes h_f) + n_{ant} + n_{Poisson} + n_{det} \quad (1)$$

where $\otimes$ denotes a two-dimensional convolution operation, $n_{ant}$ refers to antenna-induced noise, $n_{Poisson}$ represents signal-dependent Poisson noise, arising from the statistical nature of electric charge detection, $n_{det}$ encompasses other signal-independent noise source, such as thermal noise, amplifier noise and laser fluctuations. Empirically, all noise variances grow with frequency:

$$\sigma^2(f) = \sigma_0^2 + \beta f^p \quad (2)$$

where $\sigma_0^2$ is the baseline noise floor, $\beta$ and $p$ are fitted from calibration data.

TABLE I
THE PROCEDURE OF PCA-BASED THZ-SSDD ALGORITHM

**Input:** THz-TDS data $X \in \mathbb{R}^{H \times W \times B}$, PSF $h_f(x, y, f)$.
Step 1: Compute Fourier transform (fft) of $X$ and obtain spectral images;
Step 2: Feed the THz spectral images and PSF into THz-SSDD neural networks for training;
Step 3: Perform PCA decomposition for THz spectral images;
Step 4: Select the top $r$ principal components and input them into trained THz-SSDD neural networks;
Step5: Perform PCA reconstruction for the principal components processed by denoising and deconvolution.
**Output:** All reconstructed spectral images.

*B. THz-SSDD for Denoising and Deblurring*

Typical supervised learning methods train the deep neural networks (DNNs) using

$$\min_\theta \mathbb{E}_{(x,y)} \|f_\theta(y) - x\|_2^2 \quad (3)$$

where $\|\cdot\|_2^2$ denotes the squared $\ell_2$-norm loss, $f_\theta$ is a DNN with trainable parameters $\theta$, and $\mathbb{E}_{(x,y)}$ is the expectation over the joint distribution of the clean and noisy image pairs $(x, y)$. If there is no access to noise-free images, the objective function above can be re-written as

$$\min_\theta \mathbb{E}_y \|f_\theta(y) - y\|_2^2 \quad (4)$$

In this case, the DNN does not remove any noise but outputs the noisy image itself. The training Recorrupted-to-Recorrupted (R2R) scheme [24] is employed in this work to generate paired images $\{(\hat{y}, \tilde{y})\}$:

$$\begin{cases} \hat{y} = y + D^T n, \\ \tilde{y} = y - D^{-1} n, \end{cases} \quad (5)$$

where $n \sim \mathcal{N}(0, \Sigma_y)$, and $\Sigma_y = \text{diag}(H(y) - b)$. $D$ is an invertible identity matrix. $H(\cdot)$ is a $5 \times 5$ average filter that pre-smooths $y$ to estimate local variance, and $b$ denotes the background signal level. For general optical imaging systems, the unsupervised inverse problem solver can be constructed as:

$$\min_\theta \mathbb{E}_y \|\tilde{y} - (f_\theta(\hat{y}) \otimes h_f)_\downarrow\|_2^2 \quad (6)$$

where $(\cdot)_\downarrow$ denotes down-sampling operation to match the measurement grid. If the DNN is trained directly via the above objective function, it will undesirably amplify the photon noise, which will substantially contaminate the real sample information at low signal-to-noise ratio (SNR) conditions [25]. To avoid amplifying the photon noise, we proposed the terahertz self-supervised denoising and deblurring deep neural networks (THz-SSDD), which design two U-Net networks, one for denoising and the other for deconvolution. The loss function can then be given as:

$$\mathcal{L}(\theta) = \mathbb{E}_{(\hat{y},\tilde{y})} \|f_\theta(\hat{y}) - \tilde{y}\|_2^2 + \xi \mathbb{E}_{(\hat{y},\tilde{y})} \|(f_\theta(\hat{y}) \otimes h_f)_\downarrow - \tilde{y}\|_2^2 + \gamma \mathcal{R}_{Hessian}(f_\theta(\hat{y})) \quad (7)$$

where the first term enforces pixel-wise self-supervision for noise removal, the second term aligns the point spread function (PSF)-convolved output (then down-sampled) with $\tilde{y}$, driving blur inversion, $\mathcal{R}_{Hessian}(z) = \|\nabla^2 z\|_1$ penalizes spurious high-frequency artifacts, and $\xi$ and $\gamma$ balance denoising, deblurring and artifact suppression. The networks' structure is shown in Fig. 1.

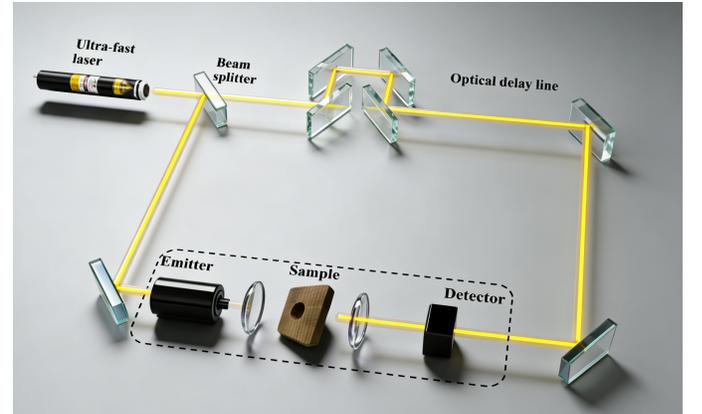

Fig. 2. The schematic image of THz-TDS systems.

*C. PSF Generation*

In this study, the point spread function (PSF) of the THz-TDS system is theoretically derived based on the diffraction-limited Gaussian beam model. This is because during the training process, the THz-SSDD can automatically optimize the model to match the input PSF kernel. For a standard optical setup employing off-axis THz lenses with focal length $f$ and aperture diameter $D$, the beam waist $\omega_0$ defined as the radius at which the intensity drops to $1/e^2$ is:

$$\omega_0 = \frac{2\lambda f}{\pi D} \quad (8)$$

where $\lambda$ is the wavelength of the THz wave corresponding to the desired frequency component. When the focusing system satisfies $\frac{f}{D} = 4$, Eq. (8) can be simplified to $\omega_0 \approx 2.547\lambda$. Based on Eq. (8), the PSF is modeled as a normalized two-dimensional Gaussian kernel:

$$h_f(x,y) = \frac{1}{2\pi\sigma_f^2} \exp\left(-\frac{x^2+y^2}{2\sigma_f^2}\right) \quad (9)$$

where $\sigma_f = \frac{\omega_0(f)}{\sqrt{2}} \approx 1.8\lambda$ is the standard deviation and scales inversely with frequency. Therefore, the 3D PSF can be calculated as:

$$h_f(x,y,f) = \frac{f^2}{6.48\pi c^2} \exp\left(-\frac{f^2(x^2+y^2)}{6.48c^2}\right) \quad (10)$$

For a 2D deep neural network, we only need a 2D PSF. Additionally, Neural networks can easily learn the features of PSF for most THz spectral images. Here, we set $f = \frac{f_i + f_e}{2}$, where $f_i$ is the selected initial frequency and $f_e$ is the selected end frequency.

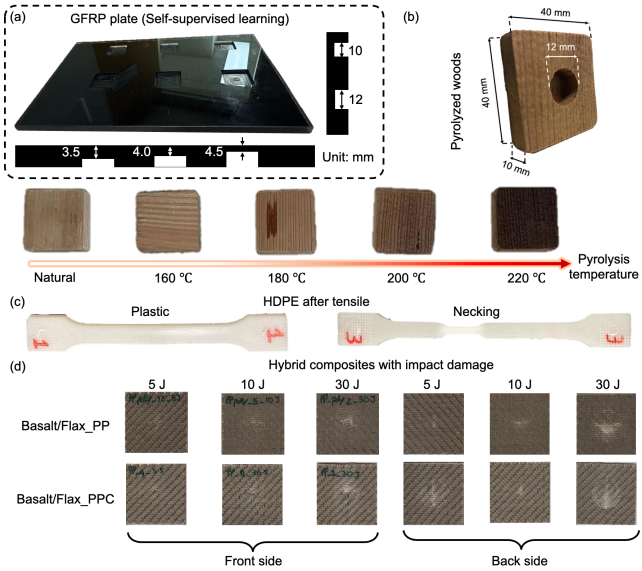

Fig. 3. **Prepared samples.** (a) GFRP plate for self-supervised learning; (b) Pyrolyzed woods under different temperature; (c) HDPE after tensile where the left HDPE undergoes plastic deformation, and right HDPE has local necking; (d) Hybrid composites Basalt/Flax_PP (up) and Basalt/Flax_PPC (down) subjected to varying impact energies.

*D. PCA-Based Image Restoration*

The THz images cannot be reconstructed if only THz-SSDD was employed. In other words, the algorithms for processing single image cannot solve the problem of low-frequency blurring and high-frequency noising. Therefore, a principle component analysis (PCA)-based image restoration strategy is proposed for THz-TDS systems.

Firstly, original spectral images $X \in \mathbb{R}^{H \times W \times B}$ (where $H$ and $W$ are spatial size, $B$ is frequency component) is transformed into a two-dimensional matrix $Y_{2D} \in \mathbb{R}^{B \times N}$ where $N = H \times W$, and each column is the spectral vector of one pixel. Then, subtract the mean of each column so that the data has zero mean:

$$\tilde{X} = X - \mu \quad (11)$$

where $\mu$ is the mean vector of each column. Now we can compute the covariance matrix:

$$C = \frac{1}{n-1}\tilde{X}^T\tilde{X} \quad (12)$$

where the covariance matrix $C \in \mathbb{R}^{B \times B}$ describes the correlation between features. The eigenvalue equation can be solved:

$$Cv = \lambda v \quad (13)$$

where $v$ is an eigenvector of the covariance matrix C (the principal component direction), and $\lambda$ is the corresponding eigenvalue (the variance along that direction). Then, we sort the eigenvalues in descending order and select the top $r$ eigenvectors $\{v_1, v_2, ..., v_r\}$ as the principal components $E \in \mathbb{R}^{B \times r}$. The original data is projected to the principal component space:

$$Z = E^T Y_{2D} \quad (14)$$

where each row of $Z$, $Z(i, :)$, represents the $i$-th principal component image.

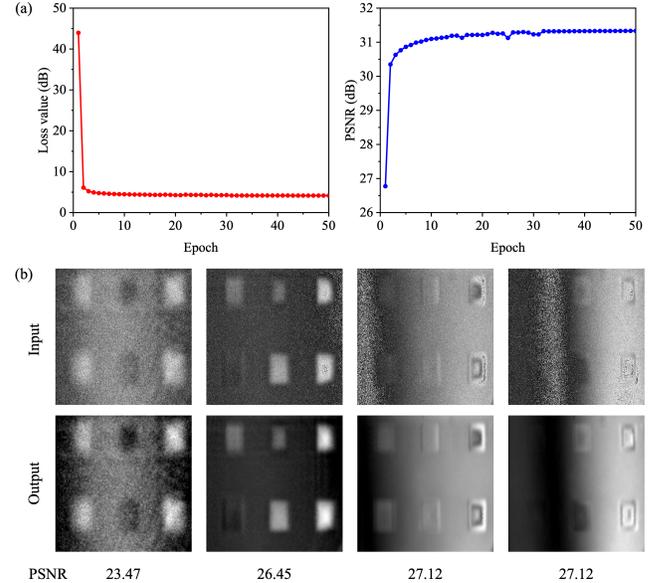

Fig. 4. **Training results for the GFRP plate.** (a) Loss value and PSNR curves during training; (b) Denoising and deblurring results for the training sample.

Now, the denoising and deconvolution neural networks, THz-SSDD, proposed in the previous section are employed to process each principal component image. Thus, the predicted images are $f_\theta(Z)$. The original data can be approximately reconstructed from the top $r$ components:

$$\hat{Y}_{2D} = E f_\theta(Z) \in \mathbb{R}^{B \times N} \quad (15)$$

Finally, reshape $\hat{Y}_{2D}$ back into the 3D cube $\hat{Y} \in \mathbb{R}^{H \times W \times B}$.

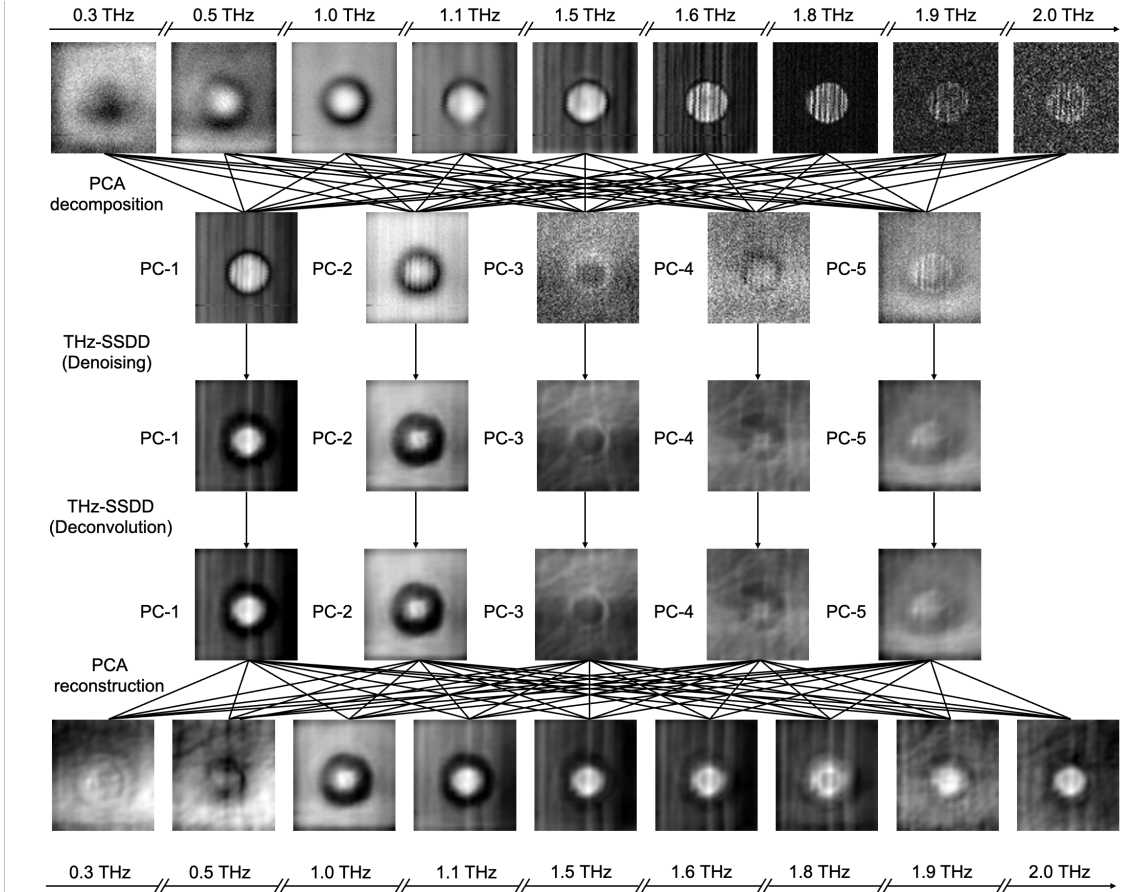

Fig. 5. **Detailed description for THz-SSDD.** The original THz-TDS results are transferred into the frequency domain based on Fourier transform. PCA decomposition was used to extract the first five principal components. Then, the trained THz-SSDD network was employed for denoising and deblurring. Finally, five processed principal components were used to reconstruct the spectral images based on PCA reconstruction algorithm.

The entire procedure is shown in Table I and Fig. 1. It is noted that too few components $r$ may cause significant information loss thus reconstructed images may lose important structural details. While too many components weaken the denoising and deblurring effect. Therefore, we should retain components that explain 95%-99% of the total variance.

### III. EXPERIMENTAL SETUP

The THz-TDS experimental set-up is shown in Fig. 2. An ultrafast laser pulse was split into a pump beam and a reference beam. The pump beam was time-shifted using an optical time-delay line and then adapted to excite a THz pulse using a THz emitter. The THz wave penetrated through the samples to a coupled detector. The reference beam was implemented on the detector as the sampling signal. The THz signal after sampling was transferred to a lock-in amplifier. The data acquisition (DAQ) system used a high-speed analog-to-digital converter (ADC) for digitizing the analog signals from the lock-in amplifier and transferring the data to a computer. Of note, the entire process could only capture single pixel. The scanning platform was used to move the sample along the horizontal and vertical directions. Repeating this process, all pixels could be obtained.

The THz system was manufactured by Menlo Systems GmbH, Munich, Germany. The experimental setup, particularly the selection of experimental parameters, is crucial in scientific research, especially in terahertz (THz) experiments. Key parameters include the operational mode, scanning step size, ambient temperature, humidity, repetition rate, and laser pulse duration. The details of experimental parameters are shown in Table II.

TABLE II
EXPERIMENTAL PARAMETERS IN THz-TDS SYSTEMS AND THz-SSDD

|  | Parameter | Value |
| --- | --- | --- |
| THz-TDS system | Mode | Transmission |
|  | Frequency resolution | 1.2 GHz |
|  | Polarization | Perpendicular to the beam direction |
|  | Scanning step | 0.5 mm |
|  | Ambient temperature | 22 °C ± 0.1 °C |
|  | Humidity | 50% ± 2% |
| THz-SSDD | Batch size | 16 |
|  | Epochs | 100 |
|  | Initial learning rate | $1\times10^{-3}$ |
|  | Optimizer | Adam |
|  | Training noise level | 25 |
|  | Weight initialization | Kaiming initialization |
|  | Training strategy | R2R |

As for the training of THz-SSDD, different from general supervised learning and unsupervised learning, a self-supervised learning strategy was employed to validate the denoising and deblurring capability of proposed THz-SSDD neural networks. The details of training are shown in Table II. The overall training time is ~4 h. The model was implemented with the PyTorch framework and then trained using NVIDIA 4060 Titan GPUs.

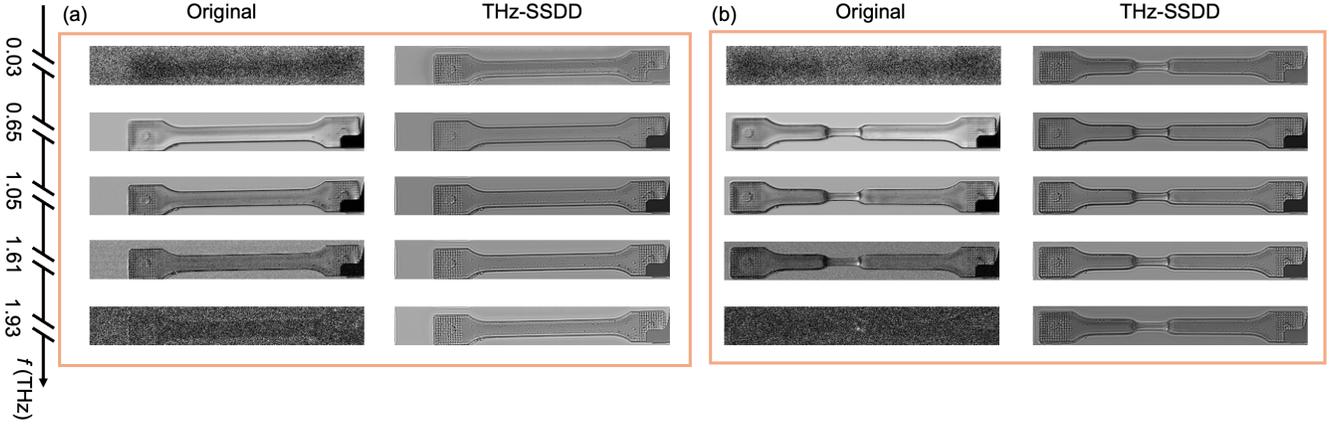

Fig. 6. **Testing results for two pure HDPE dog-bone specimens using the PCA-based THz-SSDD network.** (a) The HDPE undergoes plastic stage; (b) The HDPE undergoes necking stage.

## IV. RESULTS AND DISCUSSION

### A. Sample Preparation

To evaluate the performance of the PCA-based THz-SSDD neural networks, three types of samples were tested, as shown in Fig. 3. The first sample is a glass fibre reinforced polymer (GFRP) laminate with six flat-bottom holes. The size of upper holes is 10 mm and of the bottom holes is 12 mm. The depths of different holes are 3.5, 4.0, and 4.5 mm, respectively. The thickness of GFRP laminate is 5 mm. To demonstrate the robustness of the proposed THz-SSDD, only GFRP laminate is used for self-supervised training. The second type of samples are spruce woods with dimensions of 40 mm × 40 mm × 10 mm, as shown in Fig. 3 (b). Flat holes were made prior to the pyrolysis process. The depth of the flat holes is 5 mm, and their diameter is 12 mm. The samples were thermally modified at the temperature of 160 °C, 180 °C, 200 °C, and 220 °C, respectively. The increasing darker color is mainly caused by the loss of water during the pyrolysis process, as well as some consequent water-soluble extractions such as phenols and flavonoids, which move to the surface [17].

The third type of samples are two pure HDPE dog-bone specimens, as shown in Fig. 3(c). They were installed on the MTS® series 647 Hydraulic Wedge Grips tension machine. Each sample was damage (plastic deformation) during tensile tests. The fourth type of samples are four hybrid flax/basalt composite materials. Three samples (Basalt/Flax_PP in Fig. 3(d)) were prepared by compounding polypropylene (PP, Bormod HF955MO) with 2 wt% maleic anhydride grafted polypropylene (MA-g-PP, Polybond 3000). Three modified samples (Basalt/Flax_PPC in Fig. 3(d)) were manufactured by modified PP formulation (PPC), which used a co-rotating twin-screw extruder (Collin ZK25T) at 180-205 °C and 60 rpm. The resulting films were produced with a flat-head extruder (Collin E 20-T) and subsequently used for laminate fabrication. Specimens measuring 100 mm × 100 mm × 4 mm were subjected to low velocity impact tests at four different impact energy levels, including 5, 10, and 30 J by using an instrumented drop-weight impact testing machine (CEAST/Instron 9340) equipped with a 12.7 mm hemispherical tip and a total weight of 8.055 kg. Test coupons were pneumatically clamped between two steel plates leaving a circular unsupported area with a diameter of 40 mm.

### B. Testing Results After Self-Supervised Training

Unlike conventional super-resolution models, our method adopts a zero-shot self-supervised paradigm, where the network is optimized for randomly selected input image and acts as a reusable inference model for other tasks, as shown in Table III.

TABLE III
THE DATASET CLASSIFICATION FOR TRAINING

| Datasets | Material | Number |
|---|---|---|
| Training | GFRP | 200 |
|  | Wood | 200 |
| Testing | HDPE | 200 |
|  | Hybrid composites | 600 |

The training results are shown in Fig. 4. With the increasing of epochs, the loss value decreases while the peak signal-to-noise ratio (PSNR) increases. Where the definition of PSNR is given as:

$$PNSR = 10 \log_{10} \frac{255^2}{M \times N \sum_{i=1}^{M} \sum_{j=1}^{N} |T(i,j) - I(i,j)|^2} \quad (16)$$

where $T$ denotes the output image, $I$ denotes the input image, $M$ and $N$ represent the pixel number along length and width directions. Both loss curve and PSNR curve in Fig. 4(a) exhibit the THz-SSDD network can learn the noise feature based on the proposed Recorrupted-to-Recorrupted strategy.

To further validate the training performance of the proposed THz-SSDD framework, the training dataset was evaluated using the trained neural network, as the learning strategy adopted in this work is self-supervised. The denoising and deblurring results for the GFRP plate are shown in Fig. 4(b). Owing to system noise, the amplitude images suffer from significant degradation. As observed in the input images, severe

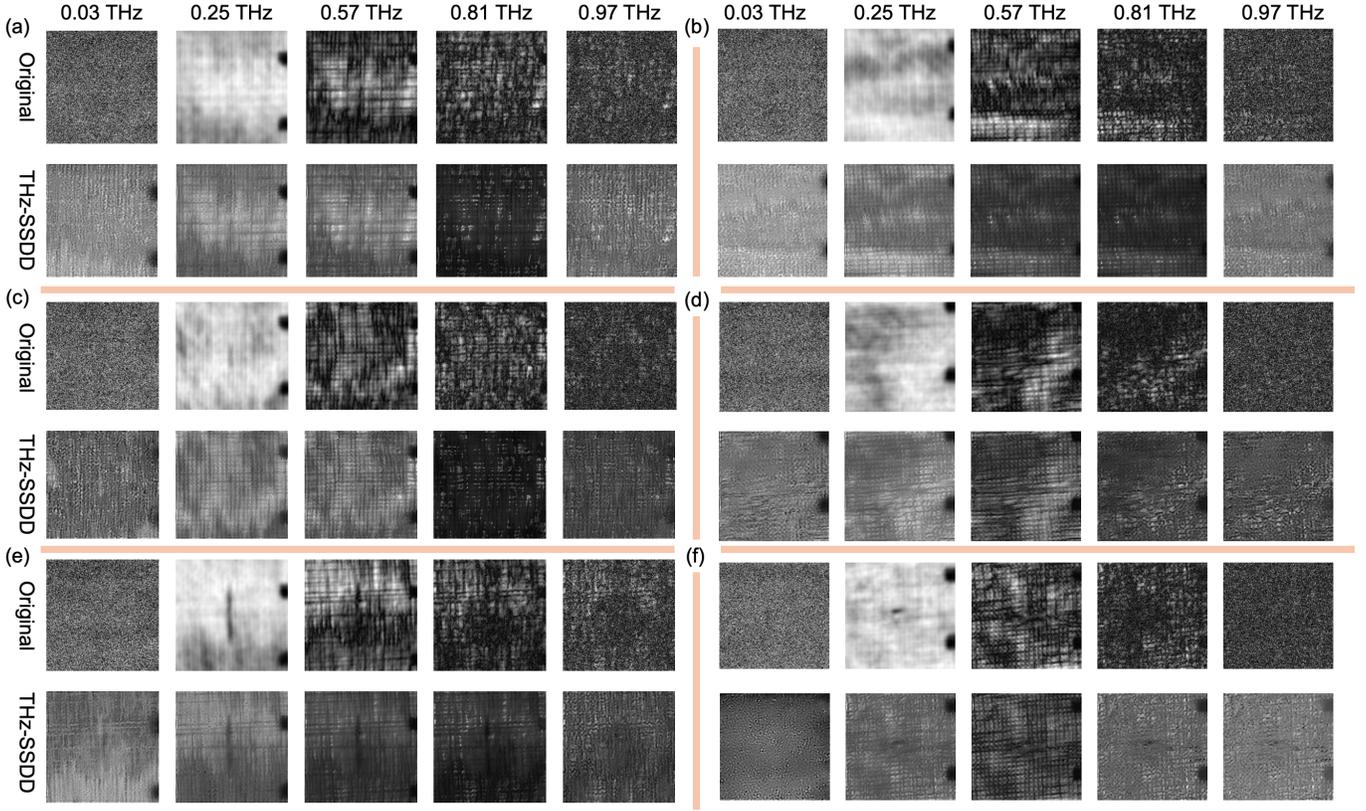

Fig. 7. **Testing results for hybrid composites Basalt/Flax_PP and Basalt/Flax_PPC.** (a), (c), and (e) are Basalt/Flax_PP under 5, 10, and 30 J impact damage. (b), (d), and (f) are Basalt/Flax_PPC under 5, 10, and 30 J impact damage.

Gaussian noise obscures the defect features and blurs the defect boundaries. After processing with the THz-SSDD network, the Gaussian noise is effectively suppressed and the defect boundaries become more distinct, as shown in Fig. 4(b), particularly in the second, third, and fourth images. PSNR is employed as a quantitative metric to assess image quality and the performance of the proposed algorithm. It can be observed that the PSNR values of the second, third, and fourth images are higher than that of the first image.

*C. Generalization Testing for Other Samples*

Although the proposed THz-SSDD network demonstrates effective denoising and deblurring performance on the training dataset, it is essential to evaluate its generalization capability. To this end, three unrelated samples were selected to assess the generalization performance of the proposed THz-SSDD network. Of note, the testing samples differ not only in their optical properties but also in the measurement configurations. Specifically, the GFRP plate was measured in reflection mode, whereas the other samples were measured in transmission mode.

To illustrate the processing pipeline of the proposed PCA-based THz-SSDD algorithm, the intermediate results at each stage are presented in Fig. 5. Due to system noise and scattering effects in the tested samples, the original amplitude images exhibit pronounced low-frequency blurring and high-frequency noise. Conventional image-processing algorithms are generally unable to address these two degradations simultaneously, as denoising and deblurring rely on fundamentally different processing strategies. Moreover, traditional approaches often require manual intervention to determine whether denoising or deblurring should be applied to a given image. In the proposed algorithm, the original amplitude images are first decomposed using PCA, and only the first five principal components (PCs) are retained. It can be observed that the informative content degrades significantly beyond the second PC. Subsequently, the proposed THz-SSDD network is applied to denoise these five PCs. The spatially overlapping noise caused by misalignment in the control system is effectively removed in the first PC, while the Gaussian noise in PC-3, PC-4, and PC-5 is substantially suppressed. After denoising, the same five PCs are further processed by the THz-SSDD network for deblurring, aiming to enhance spatial features, particularly the blurred defect boundaries. Finally, the processed PCs are employed to reconstruct the spectral information via PCA. According to the final reconstruction results in Fig. 5, the proposed PCA-based THz-SSDD framework effectively suppresses both low-frequency blurring and high-frequency noise.

Following the validation on wooden samples, two pure dog-bone specimens were further evaluated using the proposed PCA-based THz-SSDD network, as shown in Fig. 3(c). Several representative frequency components – namely 0.03, 0.65, 1.05, 1.61, and 1.93 THz - were selected to demonstrate the performance of the THz-SSDD network, as shown in Fig. 6. Similar to the results in Fig. 5, the amplitude images suffer from noticeable degradation at both low and high frequencies. After processing with the PCA-based THz-SSDD network, image details at both low and high frequencies are effectively recovered, as evidenced by the results at 0.03 and 1.93 THz in Fig. 6. In addition, pronounced background noise is observed at 1.61 THz in Fig. 6(a) and (b). The proposed PCA-based THz-SSDD network effectively suppresses this background noise across all selected frequency components.

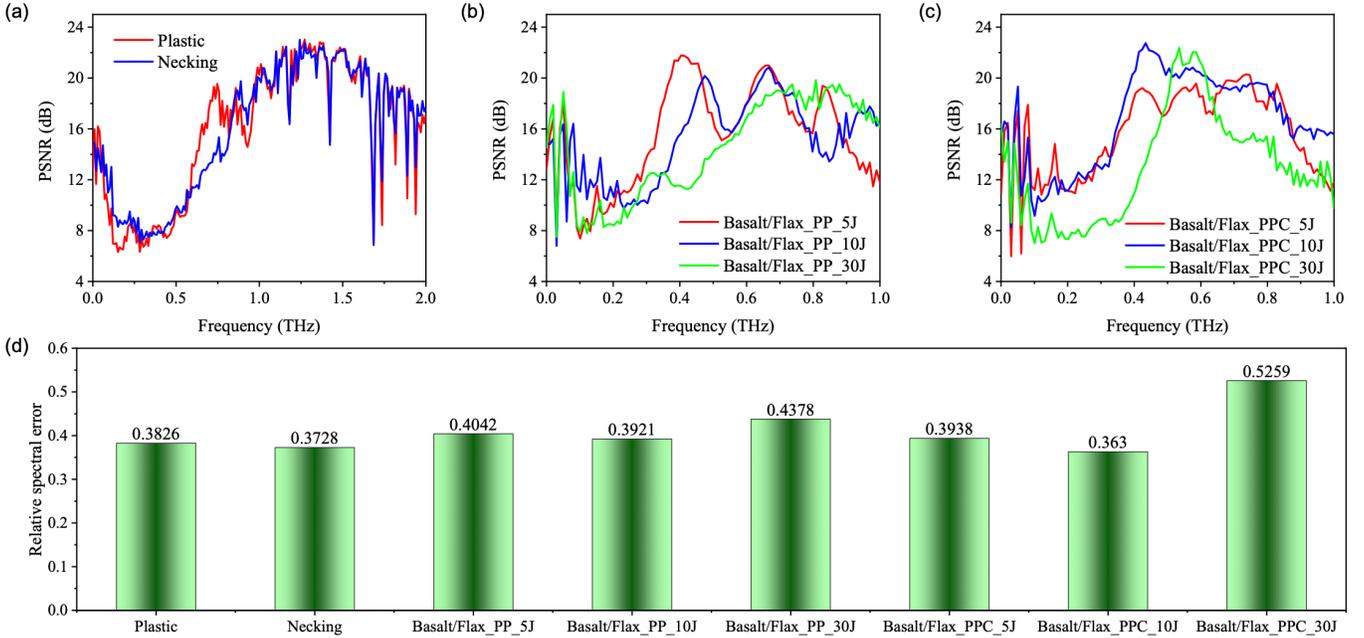

Fig. 8. **Quantitative evaluation of the proposed PCA-based THz-SSDD network.** (a) PSNR curves of HDPE samples; (b) PSNR curves of Basalt/Flax_PP samples; (c) PSNR curves of Basalt/Flax_PPC samples; (d) RSE values of HDPE samples and hybrid composites.

Finally, the PCA-based THz-SSDD network was applied to the hybrid composite materials subjected to various impact energies, as shown in Fig. 3(d). The original and reconstructed results are shown in Fig. 7. Several representative frequency components - 0.03, 0.25, 0.57, 0.81, and 0.97 THz - were selected to evaluate the performance of the THz-SSDD network. Compared with the results for wood and HDPE samples in Fig. 5 and 6, the amplitude images of the hybrid composites degrade more rapidly. This accelerated degradation is attributed to the material-specific absorption spectrum. The hybrid composites, composed of basalt and flax fibers, exhibit stronger absorption than wood and HDPE, resulting in lower signal amplitudes. In addition, the image texture of the hybrid composites is more complex than that of wood or HDPE, as shown in Fig. 7. The amplitude images contain information on impact damage, fiber patterns, and uneven resin distribution. At low frequencies, the impact damage is more prominent than the fiber patterns and resin inhomogeneity. However, at higher frequencies, the impact damage becomes increasingly obscured by other structural details, particularly at 0.97 THz.

The PCA-based THz-SSDD results are shown in Fig. 7, where panels (a), (c), and (e) correspond to the Basalt/Flax_PP samples subjected to 5, 10, and 30 J impact energies, respectively, and panels (b), (d), and (f) correspond to the Basalt/Flax_PPC samples subjected to 5, 10, and 30 J impact energies. The THz-SSDD processing effectively restores both low-frequency and high-frequency details. However, new noise is introduced at 0.81 THz. Additionally, the reconstructed fiber patterns obscure the impact damage and reduce image contrast at lower frequencies, such as 0.25 and 0.57 THz. Improvements are also observed in certain cases. For example, in the Basalt/Flax_PP and Basalt/Flax_PPC samples under 30 J impact energy, the original images at 0.57 and 0.81 THz show that the fiber patterns substantially mask the impact damage. After processing with the PCA-based THz-SSDD network, the impact damage is recovered and becomes clearly visible.

*D. Quantitative Analysis*

In the previous sections, the effectiveness of the proposed PCA-based THz-SSDD network was demonstrated and discussed. However, all results were interpreted primarily based on visual inspection, which is less persuasive in a scientific context. Therefore, in this section, quantitative metrics are introduced to further validate the denoising and deblurring performance of the proposed PCA-based THz-SSDD network.

In Section IV.B, PSNR was introduced as a widely used metric for evaluating improvements in image quality. In the context of THz-TDS, however, it is not sufficient to assess only image quality; the deviation between the original and processed signals must also be considered. This is because any image or signal processing algorithm may alter the amplitude or trend of the original signal. In general THz-TDS systems, amplitude signals are used not only to visualize defects or discontinuities but also to extract physical properties. To account for signal fidelity, we define a relative spectral error (RSE) to quantify the deviation between the original and processed signals. The RSE is formulated as follows:

$$RSE = \frac{\sqrt{\sum_{i=1}^{M}\sum_{j=1}^{N}|E_{proc}(x,y,f)-E_{orig}(x,y,f)|^2}}{\sqrt{\sum_{i=1}^{M}\sum_{j=1}^{N}|E_{orig}(x,y,f)|^2}} \quad (17)$$

where $M$ and $N$ represent the pixel number along length and width directions, $E_{proc}$ denotes the processed image, and $E_{orig}$ denotes the original image.

The PSNR curves for the HDPE samples are shown in Fig. 8(a). The PSNR values remain low in the frequency range from 0.1 to 0.6 THz. In addition, sudden drops in PSNR are observed at 1.2, 1.4, 1.6, and 1.7 THz, corresponding to the water absorption peaks in the THz spectrum. The PSNR curves for the hybrid composite samples are shown in Fig. 8(b) and (c), where similar trends are observed: PSNR values are lower in the 0.1-0.6 THz range.

The RSE values for the HDPE samples and hybrid composites are shown in Fig. 8(d). Lower RSE values indicate higher signal fidelity. As shown in Fig. 8(d), most RSE values are approximately 0.4. The Basalt/Flax_PPC sample subjected to 30 J impact exhibits the highest RSE, indicating that the signals are substantially altered during processing by the PCA- based THz-SSDD network.

## V. Conclusion

The design of the THz system inevitably introduces frequency-dependent degradation effects, which distort the amplitude images. Visually, these degradations manifest as low-frequency blurring and high-frequency noise. Conventional image-processing techniques are generally unable to address both types of degradation simultaneously. Moreover, manual intervention is often required, as the boundary between regions requiring denoising and those requiring deblurring is unknown. To overcome this challenge, we propose a PCA-based terahertz self-supervised denoising and deblurring neural network (THz-SSDD). A Recorrupted-to-Recorrupted strategy is employed for self-supervised learning, enabling the network to capture fundamental noise features by identifying invariants under the addition of paired noise. Subsequently, principal component analysis (PCA) is applied for decomposition and reconstruction to restore contaminated images across both low and high frequencies. The performance of the proposed PCA-based THz-SSDD network is evaluated using four types of samples. The network is trained using only a small set of unlabeled noisy images and is then tested on multiple samples with varying material properties and measurement configurations. The results demonstrate the network's effective denoising and deblurring capabilities. Finally, to quantitatively validate the feasibility of the THz-SSDD network, two metrics are considered: improvement in image quality and deviation between the original and processed signals. This approach ensures that the contaminated images are restored while preserving the physical characteristics of the original signals.